\pdfoutput=1

\documentclass[11pt]{article}

\usepackage{EMNLP2023}

\usepackage{times}
\usepackage{latexsym}
\usepackage{float}

\usepackage[T1]{fontenc}

\usepackage[utf8]{inputenc}

\usepackage{microtype}

\usepackage{inconsolata}
\usepackage{graphicx}
\usepackage{listings}
\usepackage{dirtytalk}
\usepackage{enumitem}
\usepackage{graphicx}
\usepackage{array}
\usepackage{booktabs}
\usepackage{xcolor}

\usepackage{{booktabs}}

\newcommand{\averitec}{AVerImaTeC}

\makeatletter
\newcommand\footnoteref[1]{\protected@xdef\@thefnmark{\ref{#1}}\@footnotemark}
\makeatother

\title{AIC CTU@AVerImaTeC: dual-retriever RAG for image-text fact checking}

\author{Herbert Ullrich \\
AI Center @ CTU FEE\\
Charles Square 13\\
Prague, Czech Republic\\
\texttt{ullriher@fel.cvut.cz} \\\And
Jan Drchal \\
AI Center @ CTU FEE\\
Charles Square 13\\
Prague, Czech Republic\\
\texttt{drchajan@fel.cvut.cz} \\}

\begin{document}
  \maketitle
\begin{abstract}
In this paper, we present our 3rd place system in the AVerImaTeC shared task, which combines our last year's retrieval-augmented generation (RAG) pipeline with a reverse image search (RIS) module.
Despite its simplicity, our system delivers competitive performance with a single multimodal LLM call per fact-check at just \$0.013 on average using GPT5.1 via OpenAI Batch API.
Our system is also easy to reproduce and tweak, consisting of only three decoupled modules -- a textual retrieval module based on similarity search, an image retrieval module based on API-accessed RIS, and a generation module using GPT5.1 -- which is why we suggest it as an accesible starting point for further experimentation.
We publish its code\footnote{\url{https://github.com/heruberuto/AVerImaTec\_Shared\_Task}} and prompts, as well as our vector stores and insights into the scheme's running costs and directions for further improvement.

\end{abstract}



\section{Introduction}
The challenge of automated fact verification has been studied extensively in previous works~\cite{10.1162/tacl_a_00454,akhtar-etal-2025-2nd,schlichtkrull-etal-2024-automated}, most commonly modelled as an NLP task with textual inputs.
With public discourse moving increasingly to social media, the task fact-checkers face, however, often goes beyond just text and language.
An important example of this phenomenon are the image-text claims, whose veracity depends not only on the textual statement itself, but also on the contents of images that come with it, whether they are authentic or edited, and whether the images are presented in the right context.

To facilitate the automation of this type of fact-checking,~\citealt{cao2025averimatecdatasetautomaticverification} publishes the \averitec{} dataset, collecting hundreds of reference image-text factchecks from human annotators, announcing the \averitec{} shared task late 2025, to establish its state of the art.

\begin{minipage}{0.9\linewidth}
    \centering
    \includegraphics[width=\linewidth]{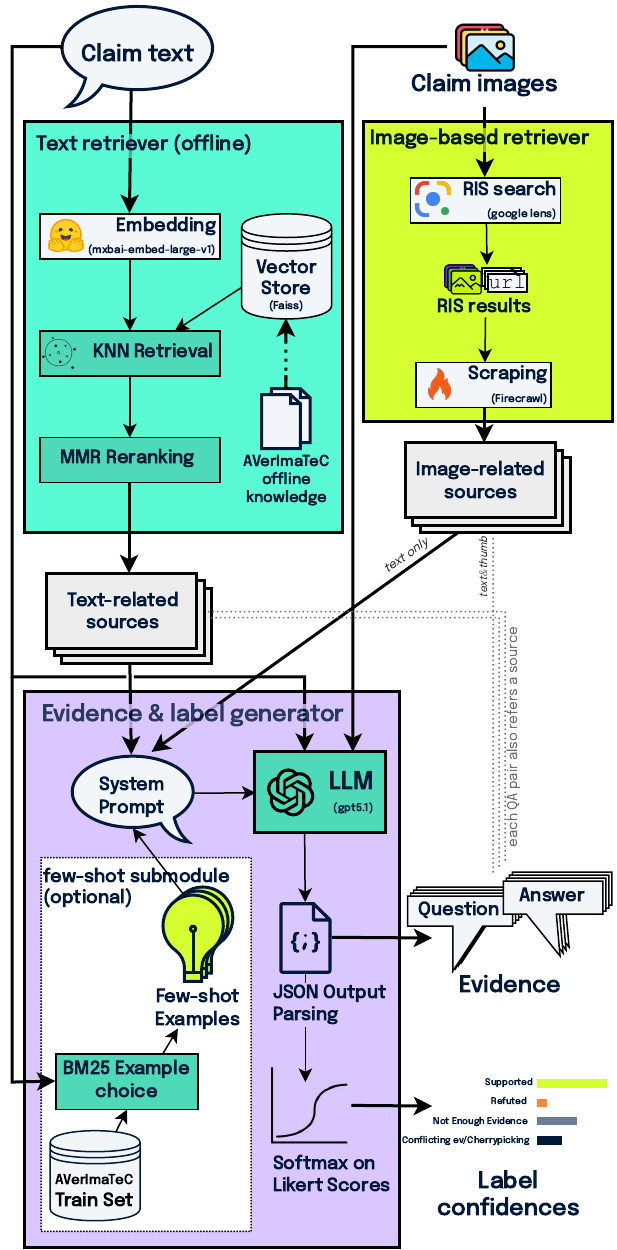}
    \captionof{figure}{Our image-text fact-checking pipeline used in CTU AIC AVerImaTeC submission, adapted from~\citealt{ullrich-drchal-2025-aic}. System is described in detail in section~\ref{sec:system2026}.}
    \label{fig:pipeline2026}
\vspace{1em}
\end{minipage}

With this paper, we introduce our 3rd place \averitec{} shared-task system, aiming to provide a strong baseline for image-text fact checking with easy-to-reimplement modules and affordable running costs. We use a single query to a multimodal LLM per claim and a single RIS request for each attached image.
Our pipeline performs retrieval-augmented generation (RAG)~\cite{rag} with two retrieval modules: one retrieves relevant documents from offline knowledge using vector search, and the other retrieves documents that contextualize the claim images using RIS (Google Lens in our case).
Our system is visualised in figure~\ref{fig:pipeline2026} and detailed in section~\ref{sec:system2026}.

\section{System description}
\label{sec:system2026}

We adapted a system from~\citealt{ullrich-drchal-2025-aic} which extends on top of~\citealt{ullrich-etal-2024-aic}.
The cited papers describe the system in detail, with ablation studies and justifications of each step.
Our pipeline, depicted in Figure~\ref{fig:pipeline2026}, is a RAG scheme of two retrievers and one generation module:

\begin{enumerate}[label=\roman*.]  
\item \textbf{Text-based retrieval module}
\begin{enumerate}[label=\arabic*.]  
    \item \textbf{Vector store} is produced for each of the \averitec{} datapoints in advance, using the scheme introduced in~\citealt{ullrich-etal-2024-aic}: the provided text-only\footnote{\averitec{} set also includes image-text and image-only knowledge stores, but since these were (as of Feb 2026) not marked with a source URL or other real-world identifier, we dropped these as inappropriate to be referred to as sources.}  \averitec{} knowledge store is chunked into 2048-character segments\footnote{The chunks do not overlap, and are annotated with context before and after in their metadata, as described in more detail in~\citealt{ullrich-etal-2024-aic}.}, and each is embedded using the \texttt{mxbai-embed-large-v1}~\cite{li-li-2024-aoe,emb2024mxbai} model.
    \item \textbf{Similarity search} is performed using the exact $k$-NN search implementation provided by the \texttt{FAISS}~\cite{douze2024faiss,johnson2019billion} library, with $k=20$ nearest neighbours
    \item \textbf{Maximal marginal relevance}~\cite{carbonell-mmr} reranking down to $l=7$ results is then applied to diversify the search results. We set the tradeoff between result diversity and similarity to the claim to $\lambda=0.8$ in favour of similarity to the claim.
\end{enumerate}
\item \textbf{Image-based retrieval module} is invoked separately for each image attached to the \averitec{} claim -- that means, if claim contains $n$ images, $n$ separate sets of results will be produced
\begin{enumerate}[label=\arabic*.]  
    \item \textbf{Reverse image search (RIS)} is performed using the \texttt{Google Lens}\footnote{\url{https://lens.google.com/}} via Serper API\footnote{\url{https://serper.dev/}} to produce a set of (up to 30) RIS results -- each assigned a webpage URL and a \textit{thumbnail}, which contains an image within this webpage similar to the given claim image -- this should be the image that triggered webpage's inclusion in RIS results.
    \item \textbf{Scraping}: each of the RIS results is then scraped, in our case, using the Firecrawl API\footnote{\url{https://firecrawl.dev/}}, which produces a LLM-friendly markdown for each of the URLs. We disregard the other images in the webpage, and only keep the thumbnail which triggered the RIS result inclusion, as it has stronger guarantees of being similar to the claim image.
    \item \textbf{Result filtering} -- to maintain the evidence principle~\cite{glockner-etal-2022-missing}, we filter out any evidence published after the claim was originally stated, using the Htmldate~\cite{barbaresi-2020-htmldate} library to estimate the publishment dates for each RIS-retrieved URL.
    
    Importantly, many results of RIS we performed in ii.1 were scraping-protected, most notably the Facebook and Instagram posts, resulting in an empty result. For simplicity and compliance with fair data usage, we toss these results as well, although we acknowledge that this might lead to a loss of useful information.

    Finally, to be able to mark the results with a single-digit identifier (iii.1) and to not clutter the generation prompt, we only preserve the first 9 of the remaining results.
\end{enumerate}
\item \textbf{Evidence, label, and justification generation module}
\begin{enumerate}[label=\arabic*.]  
    \item \textbf{System prompt} is composed of the results of both the text- (i.) and image-based (ii.) retrievers -- in the prompt, as well as in the pipeline scheme in Figure~\ref{fig:pipeline2026}, we refer to them as to \textbf{text-related sources} and \textbf{image-related sources}, respectively. We instruct the LLM to cite a source with each piece of evidence it produces, assigning the sources numerical \texttt{source ID}s: 1--9 for the text-related sources and 11--19 for the sources related to the 1st claim image, 21--29 for the sources related to the 2nd claim image, etc. 
    For the image sources, we only include their text and an information that this text was published alongside an image similar to $i$-th claim image (notably omitting the thumbnail itself), in order not to overwhelm the multimodal LLM with easy-to-confuse image inputs.
    
    These sources, as well as the task description, formatting instructions and few-shot examples (iii.2) are then serialized into a single system prompt -- its full text can be found in Appendix~\ref{appendix_sec:system2026_prompt}
    \item \textbf{Few-shot examples} of evidence are retrieved for the given claim using BM25~\cite{bm25} on \averitec{} train set. The evidence examples are then appended to the system prompt (iii.1) to make the LLM adhere better to the evidence format used by \averitec{} annotators.
    \item \textbf{Multimodal user message} is composed of the claim text in its first field, and a base64-encoded claim images in its subsequent fields.
    The user message is then passed to the LLM to generate evidence, label and justification.
    \item Upon \textbf{parsing} the LLM outputs, we augment the LLM-generated evidence which refer an image-related source with a base64-encoded \textit{thumbnail} (ii.1) of the respective image-related source to facilitatate comparison with evidence images chosen by human annotators. 
    \item \textbf{\averitec{} format matching} -- as in previous years, our system outputs the evidence formatted as QA pairs. In \averitec{}, however, this format is phased out -- while the questions are evaluated separately, the main score (see table~\ref{tab:leaderboard}) is now based on comparing two self-contained \say{evidence texts}, typically containing all the information within a single declarative sentence with pointers to relevant images.

    To match this design without introducing another LLM request, we concatenate the question and answer to obtain a self-contained evidence text for each QA pair. If an image source from RIS was referred, we append \texttt{[IMG\_1]}, referring source thumbnail, to this evidence text.
\end{enumerate}
\end{enumerate}

The system extends on top of our previous work on the AVeriTeC and AVeriTeC 2 shared tasks~\cite{ullrich-etal-2024-aic,ullrich-drchal-2025-aic}, with the most notable addition of the image-based retrieval module (ii).

%
%
%


\section{Results and analysis}
\label{nothink}

\begin{table}[h]
\centering
\begin{tabular}{l
>{\centering\arraybackslash}p{.7cm} 
>{\centering\arraybackslash}p{.7cm} 
>{\centering\arraybackslash}p{.7cm} 
>{\centering\arraybackslash}p{.7cm}}
{\small{\textbf{System}}} &
\rotatebox{70}{\textbf{\footnotesize{Question Score}}} &
\rotatebox{70}{\textbf{\footnotesize{Evidence Score}}} &
\rotatebox{70}{\textbf{\footnotesize{Verdict Accuracy}}} &
\rotatebox{70}{\textbf{\footnotesize{Justification Score}}} \\
\hline
{\small{HUMANE}}        & {0.89} & {0.54} & {0.55} & {0.56} \\
{\small{ADA-AGGR}}      & 0.37 & 0.46 & 0.54 & 0.43 \\
{\textit{\small{AIC CTU (ours)} }}       & \textit{0.81} & \textit{0.33} & \textit{0.35} & \textit{0.30} \\
{\small{XxP}}           & 0.39 & 0.27 & 0.26 & 0.20 \\
{\small{teamName}}      & 0.66 & 0.23 & 0.26 & 0.22 \\
{\small{REVEAL}}        & 0.63 & 0.28 & 0.24 & 0.13 \\
{\small{fv}}            & 0.29 & 0.16 & 0.16 & 0.13 \\
\hline
{\small{Baseline}}      & 0.55 & 0.17 & 0.11 & 0.13 \\
\end{tabular}
\caption{System leaderboard showing performance metrics on \averitec{} test-split. Our system described in section~\ref{sec:system2026} is highlighted with \textit{italics}.}
\label{tab:leaderboard}
\end{table}

The final \averitec{} leaderboard is shown in table~\ref{tab:leaderboard}. Our system achieves a combined verdict score\footnote{Proportion of claims with a correct verdict \textit{and} an evidence score of at least 0.3 at the same time, see~\citealt{cao2025averimatecdatasetautomaticverification}.} of 0.35, with a near-SOTA question score of 0.81, mean evidence score of 0.35, and a justification score of 0.3.
Metrics are based on Ev2R~\cite{akhtar2024ev2r} recall scores with LLM as a judge.

While our system does not reach the very state of the art, it significantly outperforms the iterative agentic baseline~\cite{cao2025averimatecdatasetautomaticverification} and majority of other systems across the board, scoring a solid 3rd place.
To reveal directions for future improvements, we proceed to study what its main pitfalls are using the leaderboard metrics and our own reproductions of \averitec{} dev-split metrics.

\subsection{Bottlenecks}
\label{sec:bottlenecks}
\begin{table}[h]
\centering
\begin{tabular}{l
>{\centering\arraybackslash}p{.7cm} 
>{\centering\arraybackslash}p{.7cm} 
>{\centering\arraybackslash}p{.7cm} 
>{\centering\arraybackslash}p{.7cm}}
{\small{\textbf{Evidence format}}} &
\rotatebox{70}{\textbf{\footnotesize{Question Score}}} &
\rotatebox{70}{\textbf{\footnotesize{Evidence Score}}} &
\rotatebox{70}{\textbf{\footnotesize{Verdict Accuracy}}} &
\rotatebox{70}{\textbf{\footnotesize{Justification Score}}} \\
\hline
{\small{Answer only}}   & \textbf{0.86} & 0.27 & 0.31 & 0.28 \\
{\textit{\small{Question + Answer}}}   & \textit{0.84} & \textit{0.33} & \textit{\textbf{0.39}} & \textit{0.31} \\
{\small{Declarative evidence}}   & 0.82 & \textbf{0.35} & 0.38 & \textbf{0.32} \\
\end{tabular}
\caption{Ablation study tweaking the evidence generation format from section~\ref{sec:system2026}, iii. Scheme used in final submission is in italics.}
\label{tab:ablation}
\end{table}

Looking at our standing in the leaderboard from table~\ref{tab:leaderboard}, the main bottleneck appears to be our system's \textit{evidence score}, computed using Ev2R recall. Despite the question score shows promising 81\% our lack in evidence score then propagates further to the verdict and justification scores as well.
Part of this problem could be attributed to our system's legacy evidence format geared more towards AVeriTeC 1 and 2 shared tasks -- an \textit{evidence} is generated as a QA pair, of a question fact-checker would ask themselves during the task, and an answer they would arrive to, grounded in an URL-referred source, whereas in \averitec{} evaluation scheme, the evidence is a self-contained declarative sentence with pointers to relevant images.

Table~\ref{tab:ablation} lists three approaches we took to address this discrepancy. In our first approach, we disregarded it and only listed the generated answers as \averitec{} evidence. In our second approach, which is also the one we submitted to the final leaderboard (table~\ref{tab:leaderboard}), we concatenated the question and answer to obtain each evidence string, appending a \texttt{[IMG\_1]} tag and a base64-encoded image in metadata when an image-related source was used.

To see whether this can be improved upon, we have also experimentally implemented a 3rd approach, referred to as \say{declarative evidence} in table~\ref{tab:ablation}, in which we have directly prompted the LLM to generate a self-contained declarative evidence text with pointers to used images.
Although this approach was experimental and not free of its own glitches (resulting in a malformed image pointers and \texttt{[IMG\_1]} tag being erroneously used in other generic fields, such as justification and questions), it shows promissing results, surpassing our \textit{Question+Answer} approach by encouraging 2\% in the evidence score, even before adjusting its prompt to iron out the glitches.

Another bottleneck could be possible discrepancies in our image-evidence usage -- looking closer at the ablation study in table~\ref{tab:ablation}, the \say{Answer only} approach stays too close behind its more advanced alternatives.
This finding raises concerns, since the answer-only approach does not use \textit{any} \texttt{[IMG\_1]} tags, yet per~\citealt{cao2025averimatecdatasetautomaticverification}, 53.9\% of the \averitec{} evidence should be annotated using reverse image search, with 1.6\% using the image itself as the answer. 
This is to be investigated in future works, as even a small discrepancy in the way our system presents its image sources and how the \averitec{} evaluator assumes to receive them may have a tremendous impact on the final score.

\subsection{Cost analysis}
The scheme from section~\ref{sec:system2026} uses a single RIS request per claim image (one claim may feature multiple images, but the vast majority features exactly 1 image in \averitec{}). Using Serper, this search comes at a cost of 3 credits, totalling \$0.003 with the least-discounted bulk pricing (\$50 for 50K Serper credits).

The markdown scraping was performed using the Firecrawl API, which at its hobby tier charges \$0.006 per scraped page, with 20,000 free scraping tasks for education emails. In the worst-case scenario of multiple claim images in a single claim, each with 9 RIS results older than the claim date that can be scraped\footnote{Which is not usually the case, as at least some proportion of results typically come from Meta's scraping-protected social media} and no discount, this amounts to \$0.05 per image.
To avoid this cost, however, we suggest using a free scraper instead, such as the Trafilatura library which was used to produce the \averitec{} offline knowledge stores and our system does not show any noticeable problems ingesting its outputs.

The Generation module LLM results were computed using the OpenAI Batch API, with GPT-5.1 as the backbone model.
On average, 11K completion input tokens were given to the model and 1150 tokens of output were generated per \averitec{} claim using our system from section~\ref{sec:system2026}, at an average cost of \$0.013 per claim.


\section{Conclusion}
Using a well established foundational fact-checking framework from~\cite{ullrich-etal-2024-aic,ullrich-drchal-2025-aic}, we introduce a new pipeline for image-text fact-checking using a dual-retrieval multimodal RAG system.
The two retrieval modules our system uses are a text-based similarity search and a reverse image search (RIS) accessed through an API.

Our system scores 3rd place in the \averitec{} shared task, with a combined verdict score of 0.35, a question score of 0.81, an evidence score of 0.35, and a justification score of 0.3, outperforming the baseline across the board.
With this paper, we publish a detailed description of our system design, code and prompts we used, as well as insights into the costs of its deployment and possible points of failure.
\subsection{Future works}
\label{sec:future}
\begin{enumerate}
\item During our exploratory analysis, we have witnessed many pitfalls of the used RIS engine (Google Lens) -- often providing 0 results for claims from more distant past (e.g. 2022 for dev set), or for claim images with explicit graphical content -- this should be addressed in future works, possibly swapping the RIS provider for a more robust one, as even a sub-optimal or explicit search result may be valuable for fact-checking and is better facilitated by the RAG strategy than an empty result.
\item The occasional absence of RIS results, combined with the fact that not all gold evidence includes image references, motivates an agentic extension of our pipeline (Figure~\ref{fig:pipeline2026}). An LLM controller could decide whether to use RIS, text retrieval, or both, saving resources when RIS is unlikely to help.
\item Finally, our findings in section~\ref{sec:bottlenecks} suggest possible discrepancies between how our system represents image-text evidence and how the \averitec{} evaluator expects it. Addressing this mismatch may improve scores on this benchmark and in future shared tasks derived from it.
\end{enumerate}


\section*{Limitations}
Our pipeline is not meant to be relied upon nor to replace a human fact-checker, but rather to assist an informed user. It gives sources for both the textual and image-text evidence and proposes labels for further questioning. Hallucinations may still appear in the generated justification, and the system is not meant to be used as an oracle. The current prompting and text-retrieval model assume English input, and the MLLM backbone used for the shared task (although interchangeable) is GPT5.1, which is a black box model with limited reproducibility and considerable carbon costs. Our submission also depends on proprietary services for RIS and scraping, further limiting exact reproducibility, and we do not provide a text-only vs. text+RIS ablation to isolate the RIS contribution. The cap of 9 RIS results is a prompt-budget choice that was not tuned.
Refuted class is massively overrepresented in the \averitec{} dataset (95\% of train-claims and 78\% of text-claims), making the accuracy-based \averitec{} score computed over \averitec{} test set a problematic metric for systems used in the wild.

\section*{Ethics statement}
Our pipeline extends our last-year submission. All original authors agreed with this reuse.
The system was built specifically for the~\averitec~shared task and reflects the biases of its annotators; for more information, we suggest the original \averitec{} paper~\cite{cao2025averimatecdatasetautomaticverification}.
\section*{Acknowledgements}
We would like to thank Tomáš Mlynář for providing insights into multi-modal retrieval systems.

This article was produced with the support of the Technology Agency of the Czech Republic under the SIGMA Programme, project TQ01000100 Newsroom AI: public service in the era of automated journalism.
The access to the computational infrastructure of the OP VVV funded project CZ.02.1.01/0.0/0.0/16\_019/0000765 ``Research Center for Informatics'' is also gratefully acknowledged.

\bibliography{anthology,custom}
\bibliographystyle{acl_natbib}

\appendix


\lstset{
    language={},
    basicstyle=\ttfamily\footnotesize\linespread{0.9}, 
    keywordstyle=\color{blue}\bfseries,
    commentstyle=\color{green!50!black}\itshape,
    stringstyle=\color{orange},
    numberstyle=\tiny\color{gray},
    numbers=none, 
    stepnumber=1, 
    numbersep=5pt, 
    tabsize=4, 
    showstringspaces=false, 
    breaklines=true, 
    breakatwhitespace=true,
    frame=lines, 
    captionpos=b, 
    breakindent=1em,
}
\begin{figure*}
    \section{System prompt}
    \label{appendix_sec:system2026_prompt}
    \begin{lstlisting}[breaklines=true, language={}, frame=single, caption={Our fact-checking system prompt to be used with Multimodal LLM, feeding the \averitec{} claim text and images into its multimodal user message. Three dots represent omitted repeating parts of the prompt. Adapted for multimodal scenario from~\citealt{ullrich-drchal-2025-aic}.}, label={lst:llm_system_prompt}]
You are a professional fact checker of image-text claims, formulate up to 10 questions that cover all the facts needed to validate whether the factual statement (in User message) is true, false, uncertain or a matter of opinion. The claim consists of a textual statement and {image_count} images associated with the claim. The claim was made by {author} on {date} via {medium}. Each question has one of four answer types: Boolean, Extractive, Abstractive and Unanswerable using the provided sources.
After formulating Your questions and their answers using the provided sources, You evaluate the possible veracity verdicts (Supported claim, Refuted claim, Not enough evidence, or Conflicting evidence/Cherrypicking) given your claim and evidence on a Likert scale (1 - Strongly disagree, 2 - Disagree, 3 - Neutral, 4 - Agree, 5 - Strongly agree). Ultimately, you note the single likeliest veracity verdict according to your best knowledge.
The facts must be coming from the sources listed below. The first {k} sources was retrieved using textual search and the rest was retrieved using reverse image search (google lens). The sources are numbered - sources 1 through {k} are related to the claim text,  sources 11-19 were retrieved for the first user image, 21-29 to the second etc. You may therefore assume that each of the image-based sources was published alongside a picture similar to the respective user image. 
---
## Source ID: 1 [url]
[context before]
[page content]
[context after]
...
---
## Image Source ID: 11 (related to user image 1, Title : [title], date:[page_date], url: [url], image url: [img_url])
[content]
...
---
## Output formatting
Please, you MUST only print the output in the following output format:
```json
{
 "questions":
     [
         {"question": "<Your first question>", "answer": "<The answer to the Your first question>", "source": "<Single numeric source ID backing the answer for Your first question>", "answer_type":"<The type of first answer>"},...   ],
 "claim_veracity": {
     "Supported": "<Likert-scale rating of how much You agree with the 'Supported' veracity classification>",
     "Refuted": "<Likert-scale rating of how much You agree with the 'Refuted' veracity classification>",
     "Not Enough Evidence": "<Likert-scale rating of how much You agree with the 'Not Enough Evidence' veracity classification>",
     "Conflicting Evidence/Cherrypicking": "<Likert-scale rating of how much You agree with the 'Conflicting Evidence/Cherrypicking' veracity classification>"
 },
  "veracity_verdict": "<The suggested veracity classification for the claim>",
  "verdict_justification": "<A brief justification of the veracity verdict>"
}
```
---
## Few-shot learning
You have access to the following few-shot learning examples for questions and answers.:
### Question examples for claim "{example["claim"]}" (verdict {example["gold_label"]})
"question": "{question}", "answer": "{answer}", "answer_type": "{answer_type}"
...
    \end{lstlisting}
\end{figure*}

\end{document}